\documentclass[dvipsnames]{article} %
\usepackage{colm2024_conference}

\usepackage{booktabs}
\usepackage{enumitem}
\usepackage{wrapfig}
\usepackage{algorithm}
\usepackage{algpseudocode}
\usepackage{graphicx}
\usepackage[misc]{ifsym}

\usepackage{microtype}
\usepackage{amsmath}
\usepackage{colortbl}
\usepackage[utf8]{inputenc}
\usepackage[T1]{fontenc}
\definecolor{lightgray}{rgb}{0.9,0.9,0.9}
\usepackage{caption}
\usepackage{subcaption}
\usepackage{graphicx}
\usepackage{subcaption}
\usepackage{setspace}
\usepackage{url}
\usepackage{multirow}
\usepackage{tabularx}
\usepackage{blindtext}
\usepackage{pgfplots}
\pgfplotsset{compat=1.18} 
\usepackage{tikz}
\usetikzlibrary{er,positioning,bayesnet}
\usepackage{makecell}
\usepackage{tipa}
\usepackage{siunitx}
\usepackage{nicefrac}
\usepackage{listings}
\usepackage[raster,skins, most]{tcolorbox} %
\usepackage{xltabular}
\usepackage{adjustbox}
\usepackage{xurl}
\usepackage{rotating}
\usepackage[normalem]{ulem}

\usepackage{fontawesome}

\usepackage[table]{xcolor}

\useunder{\uline}{\ul}{}

\usepackage{amsmath,amsfonts,bm}

\def\eqref#1{equation~\ref{#1}}

\def\1{\bm{1}}

\DeclareMathAlphabet{\mathsfit}{\encodingdefault}{\sfdefault}{m}{sl}
\SetMathAlphabet{\mathsfit}{bold}{\encodingdefault}{\sfdefault}{bx}{n}

\renewcommand{\ghlink}{https://github.com/Alibaba-NLP/DeepResearch}

\newcommand*\justify{%
  \fontdimen2\font=0.4em
  \fontdimen3\font=0.2em
  \fontdimen4\font=0.1em
  \fontdimen7\font=0.1em
  \hyphenchar\font=`\-
}

\renewcommand{\texttt}[1]{%
  \begingroup
  \ttfamily
  \begingroup\lccode`~=`/\lowercase{\endgroup\def~}{/\discretionary{}{}{}}%
  \begingroup\lccode`~=`[\lowercase{\endgroup\def~}{[\discretionary{}{}{}}%
  \begingroup\lccode`~=`.\lowercase{\endgroup\def~}{.\discretionary{}{}{}}%
  \catcode`/=\active\catcode`[=\active\catcode`.=\active
  \justify\scantokens{#1\noexpand}%
  \endgroup
}

\usepackage{makecell}
\usetikzlibrary{tikzmark}
\makeatletter
\newcommand*\myfontsize{%
  \@setfontsize\myfontsize{7}{8}%
}
\makeatother

\definecolor{uclablue}{RGB}{159, 195, 224}

\definecolor{uclagold}{RGB}{255, 240, 180}

\definecolor{aliceblue}{RGB}{255, 238, 241}

\definecolor{cadmiumgreen}{rgb}{0.0, 0.42, 0.24}

\definecolor{myred}{rgb}{0.7, 0.3, 0.0}
\definecolor{myblue}{rgb}{0.2, 0.3, 0.6}
\definecolor{babygreen}{rgb}{0.85, 0.97, 0.85}

\definecolor{purple1}{RGB}{126, 107, 196}
\definecolor{purple2}{RGB}{199, 158, 207}
\definecolor{purple3}{RGB}{214, 200, 255}
\definecolor{purple4}{RGB}{254, 240, 255}

\definecolor{deepblue}{RGB}{48, 58, 82}

\newcommand{\symboletongyi}{\raisebox{0pt}{~\includegraphics[scale=0.012]{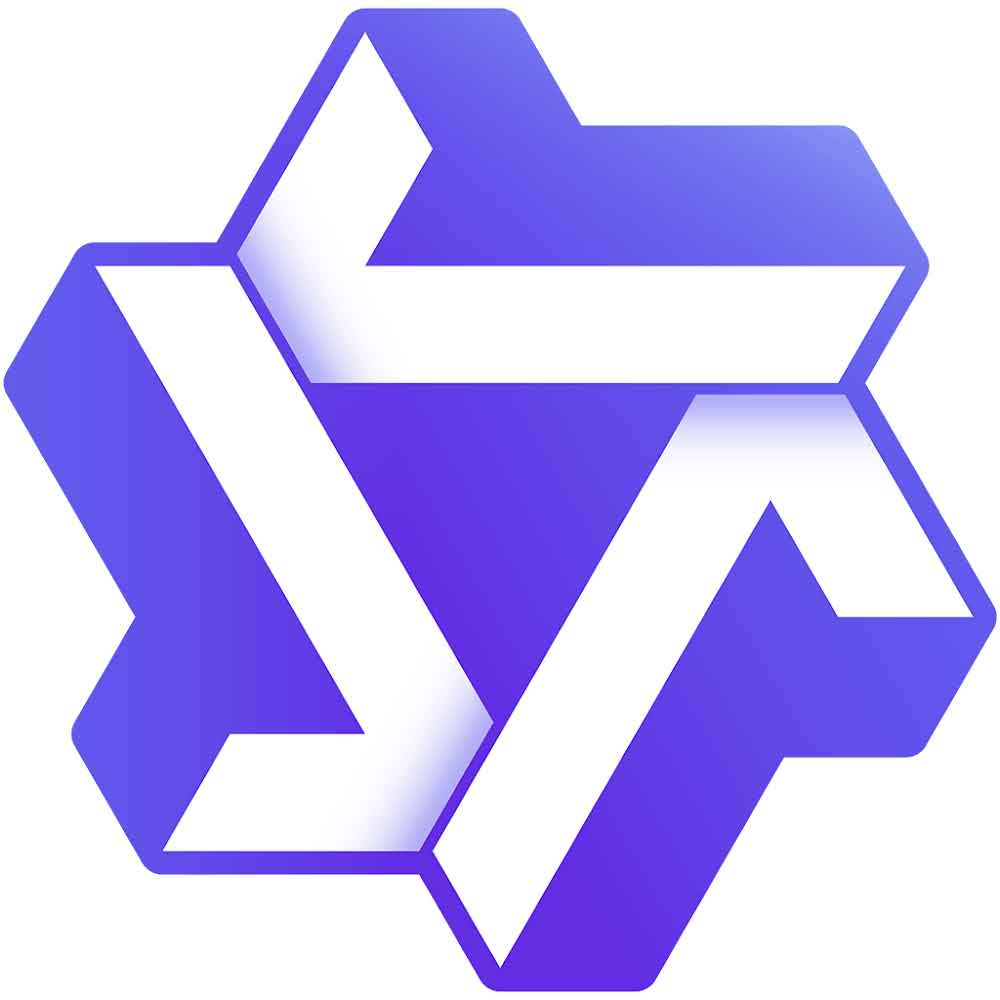}}~}

\definecolor{deepPurple}{HTML}{330066}

\definecolor{uclablue_old}{rgb}{0.15, 0.45, 0.68}
\hypersetup{
    breaklinks,
    citecolor=uclablue_old,
    colorlinks=true,
}

\newtcolorbox{mybox}[2][]
  {colback = black!5!white, colframe = black!75!black, fonttitle = \bfseries,
    colbacktitle = black!100!black, enhanced, before upper={\fontsize{8}{11}\obeyspaces\obeylines\selectfont}, fontupper=\selectfont,
    attach boxed title to top left={yshift=-2.2mm,xshift=4mm},
    title=#2,#1}

\title{
\begin{tabular}[t]{l} 
  \parbox[t]{0.8\textwidth}{\centering 
    EcomBench: Towards Holistic Evaluation of Foundation Agents in E-commerce 
  }
\end{tabular}
}

\author{%
  \small
  Rui Min\thanks{Equal contribution.},\, 
  Zile Qiao\textsuperscript{\thefootnote}\thanks{Corresponding author.},\, 
  Ze Xu,\,
  Jiawen Zhai,\,
  Wenyu Gao,\,
  Xuanzhong Chen,\,
  Haozhen Sun,\,
  Zhen Zhang,\,
  Xinyu Wang,\,
  Hong Zhou,\,
  Wenbiao Yin,\,
  Bo Zhang,\,
  Xuan Zhou,\,
  Ming Yan,\,
  Yong Jiang\footnotemark[2],\, 
  Haicheng Liu,\,
  Liang Ding\footnotemark[2],\,
  Ling Zou,\,
  Yi R. (May) Fung,\,
  Yalong Li,\,
  Pengjun Xie
  \\[1em]
  {\fontsize{10pt}{11pt}\selectfont 
  Tongyi Lab\symboletongyi, Alibaba Group}\\
}

\begin{document}

\maketitle

\begin{abstract}

Foundation agents have rapidly advanced in their ability to reason and interact with real environments, making the evaluation of their core capabilities increasingly important. While many benchmarks have been developed to assess agent performance, most concentrate on academic settings or artificially designed scenarios while overlooking the challenges that arise in real applications. To address this issue, we focus on a highly practical real-world setting, the e-commerce domain, which involves a large volume of diverse user interactions, dynamic market conditions, and tasks directly tied to real decision-making processes. To this end, we introduce \textbf{EcomBench}, a holistic \textbf{E-com}merce \textbf{Bench}mark designed to evaluate agent performance in realistic e-commerce environments. EcomBench is built from genuine user demands embedded in leading global e-commerce ecosystems and is carefully curated and annotated through human experts to ensure clarity, accuracy, and domain relevance. It covers multiple task categories within e-commerce scenarios and defines three difficulty levels that evaluate agents on key capabilities such as deep information retrieval, multi-step reasoning, and cross-source knowledge integration. By grounding evaluation in real e-commerce contexts, EcomBench provides a rigorous and dynamic testbed for measuring the practical capabilities of agents in modern e-commerce.
\end{abstract}

\begin{figure}[h]
    \centering
    \includegraphics[width=0.95\linewidth]{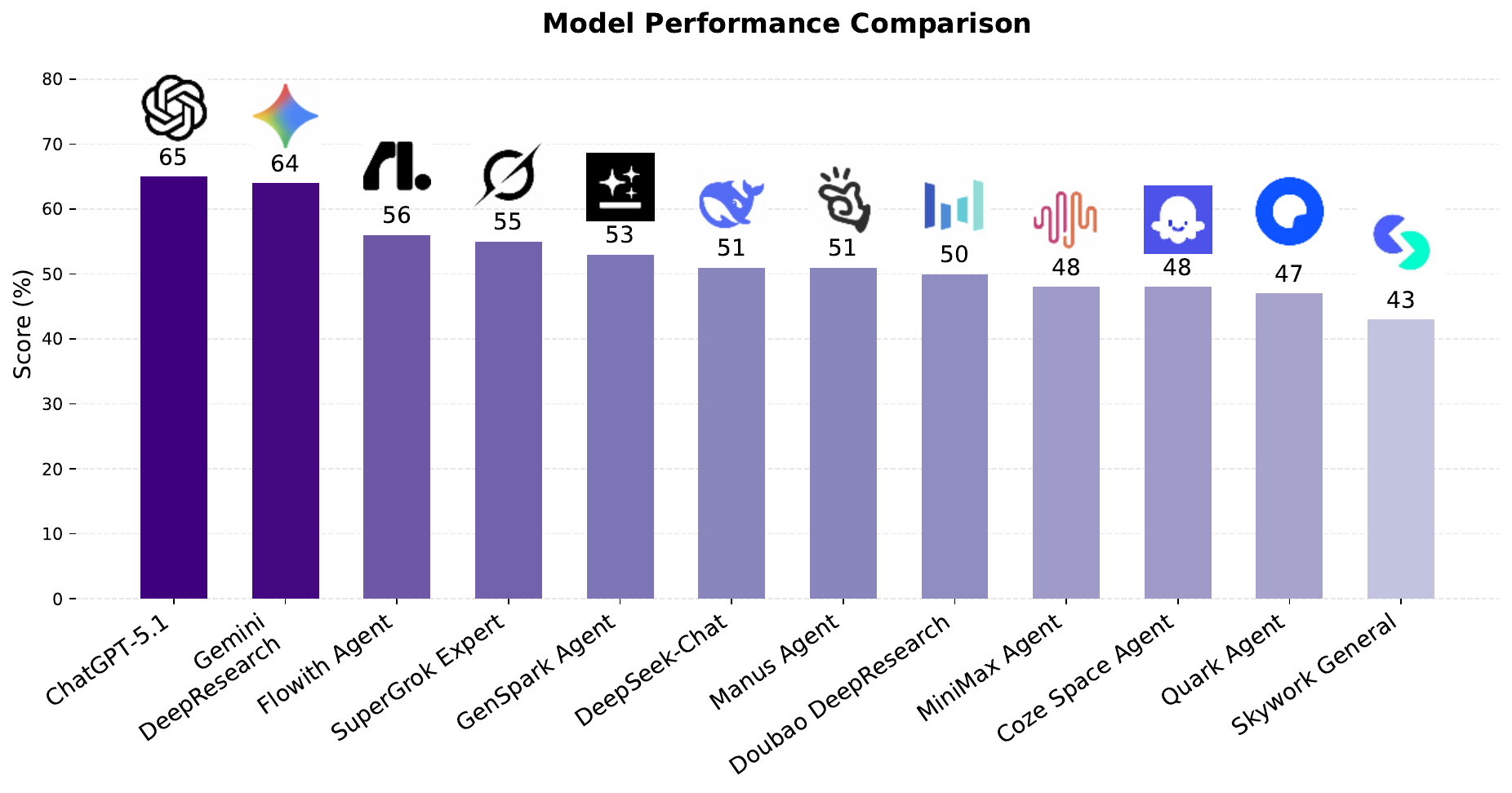}
    \caption{Comparison of model performance on the EcomBench.}
    \label{fig:abs_fig}
\end{figure}

\section{Introduction}

Large Language Models (LLMs) are rapidly advancing from passive knowledge retrievers to autonomous agents~\citep{team2025tongyi, qiu2025alita, kimiresearcher, zeng2025glm, Li2025webthinker} capable of reasoning, planning, and interacting with real-world environments. At its core, assessing the foundational capabilities of these advanced LLM agents has become essential, where a wide range of benchmarks have been proposed~\citep{mialon2023gaia,bc_en,bc_zh,xbench,yao2022webshop,wong2025widesearch,yao2024tau}. These benchmarks play a fundamental role in measuring the progress of agent development and advancing research on general agentic capabilities.

Despite the rapid growth of agent evaluation research, most existing studies focus on agent evaluation through academic puzzle-style benchmarks, leaving real-world tasks largely unexplored. E-commerce, as one of the most complex and economically significant domains, involves tasks defined by constantly changing market trends and rich domain-specific knowledge. It poses unique challenges that require agents to integrate analytical reasoning, practical expertise, and adaptive tool usage. Therefore, evaluating agents in this environment is crucial for understanding whether current models can effectively deliver relevant information and generate effective solutions within real-world e-commerce scenarios.

To rigorously assess agent capabilities in realistic e-commerce settings, we present EcomBench, a benchmark built from authentic e-commerce tasks. Unlike benchmarks that focus on environment-interaction settings~\citep{zhang2025functionality,yao2022webshop}, EcomBench centers on composite question-answering tasks that reflect authentic users’ daily issues in e-commerce. The benchmark assesses not only an agent's domain expertise but also its ability to use tools for precise information retrieval, and its capacity to capture authentic user intent and draw conclusions that genuinely benefit users. To construct EcomBench that faithfully reflects real user needs, we adopt a human-in-the-loop framework. We first collect a large volume of real-world user demands from leading global e-commerce ecosystems. From these collected data, we extract genuine user inquiries and refine them into well-defined tasks that require accurate and verifiable answers. This process yields our seed questions, which are subsequently reviewed and refined by our e-commerce experts to ensure cognitive depth, structural clarity, and answer verifiability. 

To comprehensively evaluate agents across different levels of task difficulty, we divide the benchmark into three difficulty levels and construct high-difficulty questions using a \emph{Tool Hierarchy} approach. Specifically, we employ an LLM with tools to identify hard questions: instead of relying solely on simple tools such as web search and browsing, we equip the judge model with more advanced, e-commerce-specific tools to identify questions that require multi-step, complex reasoning and cannot be solved in just a few action steps. These challenging samples demand deeper reasoning and long-horizon planning, thereby enabling a rigorous assessment of agents’ tool use, reasoning, and e-commerce domain expertise.

Empirically, we evaluate a variety of existing models on EcomBench (shown in Figure~\ref{fig:abs_fig}) and find that their performance demonstrates considerable potential for improvement in solving practical e-commerce tasks. We will regularly maintain and expand EcomBench to provide increasingly complex and realistic challenges that align with real-world e-commerce scenarios. In sum, our benchmark design follows four core principles:
\begin{itemize}
    \item \textbf{Authenticity}: EcomBench is built on large-scale genuine user demands extracted from leading global e-commerce ecosystems. These data form the basis of EcomBench, ensuring that each question reflects real e-commerce scenarios and accurately captures users’ actual needs.
    
    \item \textbf{Professionalism}: All questions are first written and refined by experienced e-commerce experts, who incorporate real user needs and domain knowledge into the question design. They are then subjected to peer validation, which ensures the clarity and overall quality of the final benchmark.

    \item \textbf{Comprehensiveness}: Our benchmark covers a wide range of e-commerce tasks, such as policy consulting, cost analysis, and marketing strategy. EcomBench also supports multiple question formats, combining multiple-choice and open-ended questions, with each question carefully calibrated for its difficulty level.
    
    \item \textbf{Dynamism}: As e-commerce continues to evolve rapidly, the benchmark is regularly updated to keep the tasks aligned with real-world market trends while reducing the potential risks of data contamination.

\end{itemize}

The rest of this paper is organized as follows. Section~\ref{sec:data_cur} illustrates \textit{Authenticity} by describing the curation process of EcomBench supported by our human-in-the-loop data engine. Section~\ref{sec:bench_analysis} provides a detailed examination of the benchmark, presenting representative examples to demonstrate its quality under \textit{Professionalism} and its fine-grained task categorization under \textit{Comprehensiveness}. To demonstrate the utility of the benchmark, Section~\ref{sec:evaluation_results} evaluates existing models on EcomBench. Finally, Section~\ref{sec:dynamic} describes the dynamic update mechanism that preserves the \textit{Dynamism} of EcomBench in the rapidly evolving e-commerce domain.
\section{Human-in-the-loop Data Curation}
\label{sec:data_cur}
The following section describes how high-quality questions are curated in EcomBench using our human-in-the-loop data engine. In Section~\ref{subsec:data_collection}, we first introduce the process of seed question collection and refinement from human experts. We then describe our question selection strategy for controlling task difficulty in Section~\ref{subsec:tool_hierarchy}.

\subsection{Data Collection from Real-World User Demands}
\label{subsec:data_collection}

E-commerce has become deeply embedded in everyday life, with millions of users generating diverse demands such as searching for policy information, estimating costs, selecting products, and making business decisions. These frequent and diverse activities provide a rich source of real-world data that captures genuine user intentions and operational needs. Building on this foundation, EcomBench leverages these authentic interactions as the basis for constructing tasks that represent real-world e-commerce scenarios. To ensure that the data reflects prevailing market trends, we derive it from real user demands embedded in leading global e-commerce ecosystems such as Amazon, while maintaining timeliness.

With the large volume of collected user demands, we transform these raw contexts into seed questions with verifiable answers. To achieve this, we prompt an LLM to examine each collected user demand and filter out samples that lack concrete answers, such as requests involving subjective evaluation criteria. These seed questions, however, are not yet ready for direct labeling and evaluation, as some may contain vague instructions that yield no ground-truth answers or are overly simple for answer retrieval. To construct valid questions for benchmarking, we engage e-commerce experts to refine them using domain-specific knowledge, ensuring that each question is well-informed by domain expertise and precisely formulated. After refinement, we conduct peer validation for answer verification, where each question is independently labeled by at least three experts. As a common practice, we discard questions with inconsistent answers among experts to ensure reliability.

\paragraph{Avoiding Purely Synthetic Questions.} After obtaining the seed questions, we primarily rely on human effort for question reconstruction and labeling, rather than using LLMs directly for question synthesis. Although this process introduces higher costs, the resulting questions more faithfully capture genuine human demands. Since e-commerce fundamentally revolves around human participation, the questions in our dataset not only reflect real-world e-commerce behaviors but also incorporate human insight and expertise of the domain.

\subsection{Question Selection with Tool Hierarchy} 
\label{subsec:tool_hierarchy}

To comprehensively evaluate agent capabilities across varying task difficulties, EcomBench supports three difficulty levels. To identify high-difficulty tasks, we adopt a tool-hierarchy-based question selection approach. Specifically, we equip the LLM with specialized e-commerce tools such as product price retrieval and trend analysis, and then apply rejection sampling to retain questions that require complex reasoning chains and cannot be solved in just a few action steps. We designate these as level-3 (high-difficulty) tasks in our dataset. Our rationale is that, for achieving the same objective, higher-level e-commerce tools can accomplish tasks in fewer steps than atomic tools such as web search or browsing, which typically require multiple actions. Consequently, these questions substantially increase the challenge for agents lacking advanced tools or e-commerce expertise, as solving them demands extensive action sequences, long-horizon reasoning, and adaptive tool usage. This filtering mechanism based on tool hierarchy provides a scalable strategy for constructing high-difficulty questions.

\section{In-depth Analysis of EcomBench}
\label{sec:bench_analysis}

\begin{table}[h]
\centering
\caption{Task categories and descriptions in EcomBench.}
\label{tab:taxonomy}
\resizebox{\textwidth}{!}{
\begin{tabular}{p{0.3\textwidth}p{0.60\textwidth}}
\toprule
\textbf{Task Category} & \textbf{Task Description} \\
\midrule
\textbf{Policy Consulting} & Tasks involving platform rules, qualification submissions, and tax registration processes, commonly seen in queries about compliance and policy-related demands in daily operations. \\
\midrule
\textbf{Cost and Pricing} & Tasks related to checking order profit, preparing quotes, and adjusting prices under different market or customer conditions, often raised when users assess profitability. \\
\midrule
\textbf{Fulfillment Execution} & Tasks covering shipping arrangements, handling returns and exchanges, and improving basic logistics routes, frequently asked about in day-to-day fulfillment issues. \\
\midrule
\textbf{Marketing Strategy} & Tasks involving planning promotions, setting up ads, and finding ways to reach users, typically appearing in queries about boosting traffic or visibility. \\
\midrule
\textbf{Intelligent Product Selection} & Tasks using trend signals and simple data insights to identify product categories with good potential, reflected in many questions about choosing the right products to sell. \\
\midrule
\textbf{Opportunity Discovery} & Tasks looking at data to spot early signs of new opportunities, often asked when users explore new directions for growth. \\
\midrule
\textbf{Inventory Control} & Tasks involving safety-stock planning, restocking decisions, and clearance actions, commonly seen in questions about balancing stock availability and overstock risks. \\
\bottomrule
\end{tabular}
}
\end{table}

\subsection{Fine-Grained Task Category}
To ensure a comprehensive evaluation across diverse user demands, EcomBench encompasses a broad spectrum of task types, primarily covering seven categories commonly observed in real-world e-commerce scenarios: Policy Consulting, Cost and Pricing, Fulfillment Execution, Marketing Strategy, Intelligent Product Selection, Opportunity Discovery, and Inventory Control. We provide a detailed description in Table~\ref{tab:taxonomy}.

\begin{figure}[htbp]
\centering
\caption{Examples of questions with different difficulty levels from \textbf{EcomBench}.}
\label{fig:level_examples}
\framebox{%
\begin{minipage}{15cm}
\ttfamily\footnotesize
{\color{blue}Level 1 (Policy Consulting):}\\
Our company has developed a new laptop power adapter with a nameplate output power of 48 watts, and we plan to sell it nationwide in the United States in 2025. To meet U.S. energy regulations, what is the maximum no-load power consumption allowed for this adapter? Please round the result to two decimal places.
\\ \\
{\color{purple}Answer:}\\
\textbf{0.1 W} \\ \\

{\color{blue}Level 1 (Cost and Pricing):}\\
Given that the Spanish home decor e-commerce market is projected to have a compound annual growth rate of 6.8\% from 2025 to 2028, and that Mediterranean-style decor is expected to grow 2.2 percentage points faster than the overall home decor category, what is the cumulative growth rate of Mediterranean-style decor over these three years? Round the result to two decimal places.\\ \\
{\color{purple}Answer:}\\
\textbf{29.50\%} \\
\textcolor{brown}{\rule{\linewidth}{0.4pt}} \\

{\color{blue}Level 2 (Policy Consulting):}\\
According to Canadian toy safety regulations, a batch of 10,000 plastic toys intended for children under three years old is inspected using an AQL 1.0 sampling standard. If 2\% of the toys actually exceed the lead limit of 90 mg/kg, what is the probability (in percent) that the batch will be incorrectly accepted? Round to the nearest whole number. \\ \\
{\color{purple}Answer:}\\
\textbf{79\%} \\ \\

{\color{blue}Level 2 (Cost and Pricing):}\\
A UK company sells a customized product bundle to consumers in Germany, consisting of: electronics priced at £200 (standard VAT rate), physical books priced at £50 (reduced VAT rate), and a digital course priced at £100 (subject to the digital services VAT rate). Given: exchange rate £1 = €1.18; German VAT rate for digital services is 9\%; EU import rules state that if the total value of the goods exceeds €150, customs duties apply (5\% for electronics, 0\% for books); customs duties are calculated based on the value of the goods (digital services excluded); VAT is calculated based on the value of goods plus customs duties plus the value of digital services; and customized electronics require a 15\% configuration fee (applies only to the electronics). Calculate the total amount (in euros) the German consumer must pay, and round the result to two decimal places.\\ \\
{\color{purple}Answer:}\\
\textbf{530.86 euros} \\
\textcolor{brown}{\rule{\linewidth}{0.4pt}} \\

{\color{blue}Level 3 (Policy Consulting):}\\
For a laptop power adapter with a nameplate output power of 48 watts that complies with the U.S. Department of Energy (DOE) Level VI efficiency standard, what is the minimum allowable input power, in watts, when the adapter delivers its full 48-watt output? Please round your result to four decimal places. \\ \\
{\color{purple}Answer:}\\
\textbf{55.1724 W} \\ \\

{\color{blue}Level 3 (Cost and Pricing):}\\
An intelligent doorbell operates in the 5.8 GHz band (5820 MHz) under the EU Radio Equipment Directive (RED). Its transmitter output power is 500 mW, using a 4 dBi antenna, with a cable loss of 1.5 dB. Calculate the equivalent isotropically radiated power (EIRP) in dBm. If the device uses an 800 kHz modulation bandwidth, determine the minimum required out-of-band emission attenuation according to EN 300 328. Finally, verify whether this configuration complies with Article 3(2) of the RED for short-range devices. Output the EIRP and attenuation as integers, and the compliance result as “Yes” or “No”. Round all results to the nearest integer.\\ \\
{\color{purple}Answer:}\\
\textbf{30, 50, No}

\end{minipage}
}

\end{figure}

\subsection{Case Study on Difficulty Levels}
In addition to the diversity of e-commerce task types, our benchmark includes tasks of different difficulty levels, which are designated as follows: We manually annotate and categorize each task into three levels. 
\textbf{Level 1 (20\%)} includes relatively simple cases that evaluate an agent’s foundational expertise in e-commerce and its ability to perform basic tool operations. \textbf{Level 2 (30\%)} contains moderately complex tasks that require the agent to decompose problems and reason across multiple action steps to reach a solution. \textbf{Level 3 (50\%)} consists of the most challenging tasks in our dataset, which are both manually verified and constructed through the tool-hierarchy process. These tasks involve cross-source knowledge integration, deep information retrieval, and long-horizon reasoning and planning capabilities.

For clarity, we sample questions from the \textit{Policy Consulting} and \textit{Cost and Pricing} categories across three difficulty levels and present these examples in Figure~\ref{fig:level_examples}. These examples illustrate that our benchmark is not limited to simple information retrieval but instead evaluates agents across multiple dimensions, requiring them to gather information from diverse sources, interpret domain-specific regulations, and make coherent decisions within realistic scenarios.

\section{On Evaluating Diverse Agents on EcomBench}
\label{sec:evaluation_results}
\subsection{Evaluation Settings}
\paragraph{Backbones} To comprehensively assess existing model performance on EcomBench, we evaluate a diverse set of leading models, including Doubao DeepResearch~\citep{doubao}, Quark Agent, DeepSeek-Chat~\citep{deepseekv3.1}, ChatGPT-5.1~\citep{gpt5}, Gemini DeepResearch~\citep{geminiresearch}, MiniMax Agent~\citep{minimaxagent}, SuperGrok Expert~\citep{grok4}, Flowith Agent~\citep{flowithagent}, Skywork General~\citep{skywork}, Manus Agent~\citep{manus}, GenSpark Agent~\citep{gensparkagent}, and Coze Space Agent~\citep{cozeagent}.


\paragraph{Scoring Methodology} While each question in EcomBench has a uniquely verifiable ground-truth answer, model outputs may contain semantically equivalent responses expressed in different forms. To ensure a fair and precise evaluation, we prompt an LLM to compare each response with its corresponding ground-truth answer. For each question, the judge assigns a binary score of 1 if the response is correct and 0 otherwise, and we report the average correctness over all questions for each model. We also manually inspect a subset of the evaluations to verify the consistency of the automatic scoring.

\subsection{Evaluation across Difficulty Levels}
\begin{figure}[h]
    \centering
    \includegraphics[width=0.95\linewidth]{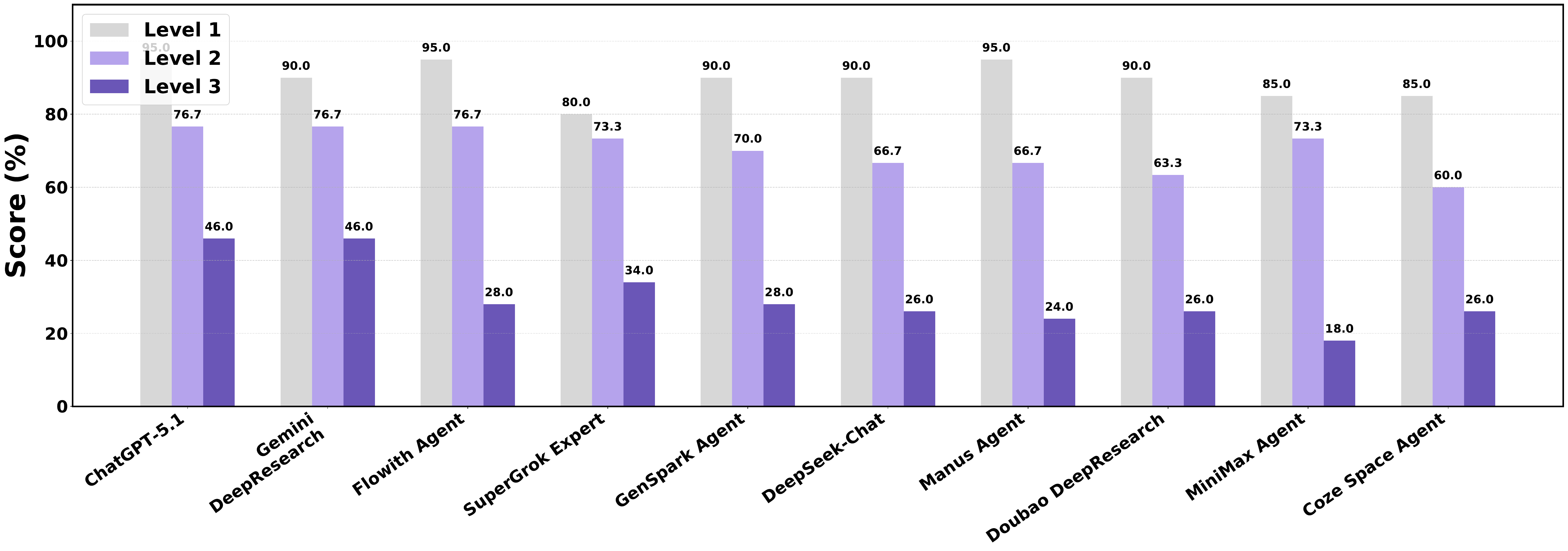}
    \caption{Performance comparison across different difficulty levels in EcomBench.}
    \label{fig:exp_difficulty}
\end{figure}

To validate our difficulty stratification, we evaluate the top ten models across three levels. As shown in Figure~\ref{fig:exp_difficulty}, we observe a progressive performance decline as difficulty increases, validating the effectiveness of our design. Specifically, models exhibit high proficiency in \emph{Level 1}, with scores consistently ranging between 80\% and 95\%. However, the performance declines noticeably in \emph{Level 2} and drops sharply in \emph{Level 3}. Even the leading models, ChatGPT-5.1 and Gemini DeepResearch, which achieve over 90\% accuracy in Level 1, see their scores fall to 46\% in \emph{Level 3}. For the remaining assessed models, the scores of \emph{Level 3} generally remain below 35\%. This substantial gap confirms that while current flagship models can reliably handle basic tasks in daily e-commerce scenarios, high-difficulty requests involving complex constraints remain a significant challenge.

\subsection{Evaluation across Task Categories}

\begin{figure}[h]
    \centering
    \includegraphics[width=0.95\linewidth]{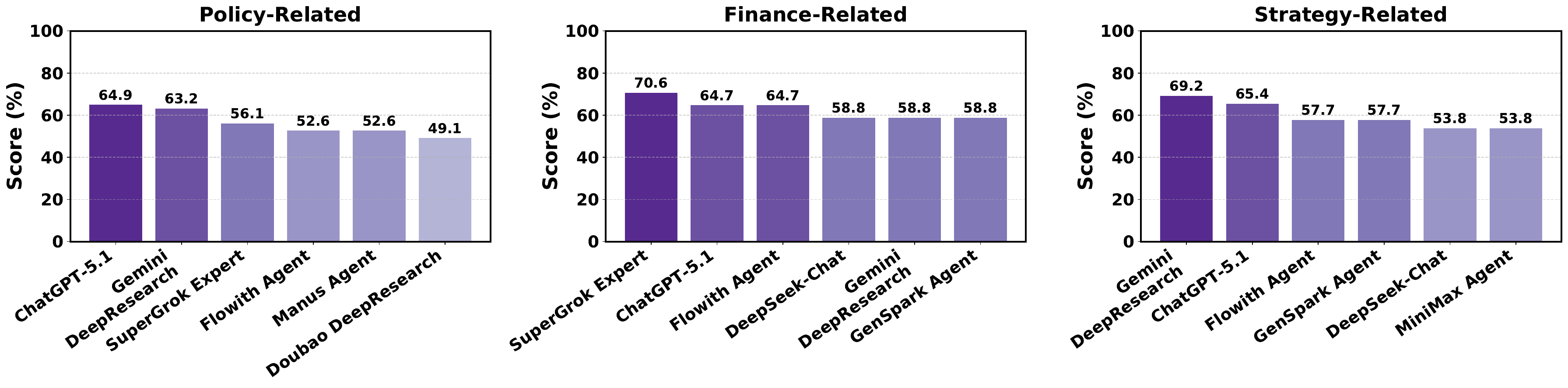}
    \caption{Performance comparison across different task categories in EcomBench.}
    \label{fig:exp_category}
\end{figure}

To further investigate the performance of distinct models across various task categories, we organize the task categories into three domains for clearer presentation: Policy-Related (comprising Policy Consulting and Fulfillment Execution), Finance-Related (comprising Cost and Pricing and Inventory Control), and Strategy-Related (comprising Opportunity Discovery, Intelligent Product Selection, and Marketing Strategy). Our objective is to examine how models perform in specialized e-commerce contexts and identify potential variations in their domain expertise. As illustrated in Figure~\ref{fig:exp_category}, which presents the top six models for each category, we observe clear performance disparities. Each model demonstrates distinct strengths; for instance, SuperGrok excels in Finance-Related tasks but falls short in Strategy-Related scenarios. Additionally, while ChatGPT-5.1 leads the overall leaderboard, it is surpassed by SuperGrok and Gemini DeepResearch in the Finance-Related and Strategy-Related categories, respectively, indicating that different models exhibit domain-specific strengths. This poses a challenge for building a more general e-commerce agent.

\section{Dynamic Maintenance and Updating of EcomBench}
\label{sec:dynamic}
\paragraph{Quarterly Update Cycle}
EcomBench is updated on a quarterly basis to ensure that it remains timely and coherent with real e-commerce scenarios. Each update serves two main purposes. First, as foundation agents continue to improve, many existing questions may no longer pose meaningful challenges. These overly simple items are replaced with new tasks that require more complex reasoning and more adaptive tool usage, thereby keeping the overall difficulty aligned with the progress of contemporary models. Second, the e-commerce domain continues to evolve, such as policy revisions, market fluctuations, and emerging product trends, causing some existing questions to become outdated. To address this, each quarterly release introduces updated questions that reflect the latest developments while removing questions that no longer match current practice. Through this continuous renewal, EcomBench remains both challenging for advancing agents and consistent with real-world e-commerce dynamics.

\paragraph{Ever-expanding E-commerce Tasks}

The initial release of EcomBench includes questions with verifiable and concise answers that cover common e-commerce scenarios. We plan to expand the benchmark with additional task types that capture a wider range of real-world e-commerce challenges, such as market analysis and forecasting. This expansion aims to move beyond simple fact-based questions toward more analytical, decision-oriented, and predictive tasks. As the benchmark evolves, it will better capture the complexity of e-commerce practice and offer more discriminative evaluations of agent capabilities.
\section{Related Work}

\paragraph{DeepResearch Agent for Complex Information Seeking} The development of deep research agents stems from foundational advances in retrieval augmentation, agentic frameworks, and training methodologies. The effort to overcome the static knowledge limitations of LLMs began with Retrieval-Augmented Generation (RAG)~\citep{lewis2020retrieval}, which integrates external information. Early single-step methods evolved into more sophisticated, multi-step workflows featuring iterative retrieval~\citep{shao2023enhancing}, query planning~\citep{ma2023query}, and self-critique mechanisms~\citep{asai2024self}. However, these RAG pipelines often lack the dynamic, autonomous decision-making characteristic of true agentic systems. A paradigm shift toward autonomy was marked by agentic frameworks like ReAct~\citep{yao2023react}, which synergizes reasoning (thought) and acting (tool use) in an interleaved manner. This model forms the conceptual backbone for modern deep research agents, including WebThinker~\citep{Li2025webthinker}, WebDancer~\citep{wu2025webdancer}, and Tongyi DeepResearch~\citep{qiao2025webresearcher, team2025tongyi}, which extend the agentic paradigm to handle long-horizon, complex information-seeking tasks through specialized tools and structured reasoning processes. The scarcity of high-quality training data for such complex tasks has driven a focus on automated synthesis. While initial information-driven methods generated questions from pre-crawled web content~\citep{wu2025webdancer}, they could suffer from structural inconsistencies. To address this, more structured, formalization-driven paradigms have emerged. These include using set-theoretic constructs to ensure consistency~\citep{tao2025webshaper} or employing tool-augmented complexity escalation to systematically generate verifiable, super-human level datasets~\citep{li2025websailor, li2025websailorv2, qiao2025webresearcher, team2025tongyi}.

\paragraph{Evaluation of Foundation Agents} The evaluation of LLM agents has rapidly evolved from assessing static knowledge retrieval, established by RAG frameworks~\citep{lewis2020retrieval} and early benchmarks like HotpotQA~\citep{yang2018hotpotqa}, to measuring dynamic, real-world interaction. A pivotal shift was marked by GAIA~\citep{mialon2023gaia}, which introduced multi-step, tool-dependent tasks with verifiable factual answers, establishing a robust paradigm for evaluating generalist assistants. As model capabilities advanced, subsequent benchmarks have diverged to probe the frontiers of agent intelligence. One direction tests profound, "Google-proof" expert reasoning, exemplified by GPQA~\citep{rein2024gpqa} and Humanity's Last Exam~\citep{phan2025humanity}. Another direction examines procedural reliability in complex and noisy environments. This includes testing an agent's ability for persistent traversal of entangled information~\citep{bc_en}, deep vertical exploration of websites~\citep{wu2025webwalker}, reasoning under conflicting search results~\citep{pham2025sealqa}, and navigating the unique challenges of non-English ecosystems~\cite{bc_zh}. More recently, the focus has expanded beyond technical performance to measure economic impact, a trend pioneered by xbench~\citep{xbench}.

Recently, the research community has shifted its focus toward evaluating agents in more realistic scenarios. For instance, FutureX~\citep{zeng2025futurex} investigates agents’ ability to perform future prediction tasks; FinSearchComp~\citep{hu2025finsearchcomp} assesses their capabilities in financial information seeking, and StockBench~\cite{chen2025stockbench} evaluates agents within realistic stock trading environments. Along this line, ECom-Bench~\citep{wang2025ecom} evaluates multimodal agents through persona-based user simulations, while our EcomBench focuses on more general e-commerce expertise, covering a broader range of domains and decision-driven tasks grounded in practical e-commerce scenarios.

\section{Conclusion and Limitations}
In this paper, we present EcomBench, a holistic benchmark for evaluating agent capabilities in realistic and dynamic e-commerce scenarios. We focus on the e-commerce domain, which represents a complex and rapidly evolving real-world environment characterized by diverse user interactions and high-stakes decision-making. EcomBench is built on four core principles: \emph{authenticity}, derived from genuine user demands in global e-commerce ecosystems; \emph{professionalism}, ensured through a human-in-the-loop curation process involving domain experts; \emph{comprehensiveness}, encompassing multiple task categories and three distinct difficulty levels; and \emph{dynamism}, maintained through a quarterly update cycle that reflects market trends and mitigates data contamination. These principles make EcomBench a rigorous and realistic testbed for evaluating the practical capabilities of foundation agents. 

\paragraph{Limitations.} EcomBench currently focuses on question-answering tasks and does not explicitly evaluate agents in environments with interactions. In addition, many real-world e-commerce problems are inherently predictive, such as product selection and market trend forecasting, which we will incorporate in future releases. As e-commerce is a highly dynamic and human-centric domain, maintaining the overall quality of the benchmark requires substantial human effort. Consequently, long-term maintenance depends on regular problem design and verification, which inevitably increases the cost of dataset construction. Despite these limitations, EcomBench offers a valuable testbed, and we view it as a living benchmark that we will continuously refine and expand to enable more comprehensive evaluation in real e-commerce scenarios.

\clearpage
\appendix

\clearpage
\bibliography{biblio}
\bibliographystyle{colm2024_conference}

\end{document}